\documentclass[sigconf,nonacm]{acmart}

\AtBeginDocument{%
  }

\usepackage{amsmath,amssymb,amsfonts}
\usepackage{algorithmic}
\usepackage{algorithm}
\usepackage{graphicx}
\usepackage{textcomp}
\usepackage{tikz}
\usetikzlibrary{arrows.meta,positioning,fit,calc}

\usepackage[table]{xcolor}
\usepackage{booktabs}
\usepackage{academicons}
\usepackage{ragged2e}
\usepackage{url}
\definecolor{orcidlogocol}{HTML}{A6CE39}
\usepackage{multirow}
\usepackage{tabularx}
\usepackage{makecell}
\usepackage{tabularx}
\usepackage{subcaption}
\usepackage{comment}
\usepackage{array}
\usepackage{threeparttable}

\usepackage{colortbl} 
\usepackage{xcolor}   

\definecolor{lightgreen}{RGB}{220, 255, 220}

\usepackage{booktabs} 
\usepackage{multirow}

\definecolor{cInput}{RGB}{235,245,255}
\definecolor{cEnc}{RGB}{237,250,241}
\definecolor{cLatent}{RGB}{255,244,235}
\definecolor{cOut}{RGB}{245,240,255}

\definecolor{cBefore}{RGB}{235,245,255} 
\definecolor{cEdit}{RGB}{237,250,241}   
\definecolor{cAfter}{RGB}{255,244,235}  

\definecolor{cHighlight}{HTML}{EAF7EE}

\begin{document}

\title{CARE-ECG: Causal Agent-based Reasoning for Explainable and Counterfactual ECG Interpretation}


\author{Elahe Khatibi}
\authornote{Equal contribution.}
\affiliation{%
  \institution{University of California, Irvine}
  \city{Irvine}
  \state{California}
  \country{United States}
}
\email{ekhatibi@uci.edu}

\author{Ziyu Wang}
\authornotemark[1]
\affiliation{%
  \institution{University of California, Irvine}
  \city{Irvine}
  \state{California}
  \country{United States}
}
\email{ziyuw31@uci.edu}

\author{Ankita Sharma}
\affiliation{%
  \institution{Arizona State University}
  \city{Tempe}
  \state{Arizona}
  \country{United States}
}
\email{ashar236@asu.edu}

\author{Krishnendu Chakrabarty}
\affiliation{%
  \institution{Arizona State University}
  \city{Tempe}
  \state{Arizona}
  \country{United States}
}
\email{Krishnendu.Chakrabarty@asu.edu}

\author{Sanaz Rahimi Moosavi}
\affiliation{%
  \institution{California State University, Dominguez Hills}
  \city{Carson}
  \state{California}
  \country{United States}
}
\email{srahimimoosavi@csudh.edu}

\author{Farshad Firouzi}
\affiliation{%
  \institution{Arizona State University}
  \city{Tempe}
  \state{Arizona}
  \country{United States}
}
\email{Farshad.Firouzi@asu.edu}

\author{Amir Rahmani}
\affiliation{%
  \institution{University of California, Irvine}
  \city{Irvine}
  \state{California}
  \country{United States}
}
\email{a.rahmani@uci.edu}

\renewcommand{\shortauthors}{Khatibi and Wang et al.}

\begin{abstract}
Large language models (LLMs) enable waveform-to-text ECG interpretation and interactive clinical questioning, yet most ECG-LLM systems still rely on weak signal--text alignment and retrieval without explicit physiological or causal structure. This limits grounding, temporal reasoning, and counterfactual “what-if” analysis central to clinical decision-making. We propose CARE-ECG, a causally structured ECG-language reasoning framework that unifies representation learning, diagnosis, and explanation in a single pipeline. CARE-ECG encodes multi-lead ECGs into temporally organized latent biomarkers, performs causal graph inference for probabilistic diagnosis, and supports counterfactual assessment via structural causal models. To improve faithfulness, CARE-ECG grounds language outputs through causal retrieval-augmented generation and a modular agentic pipeline that integrates history, diagnosis, and response with verification. Across multiple ECG benchmarks and expert QA settings, CARE-ECG improves diagnostic accuracy and explanation faithfulness while reducing hallucinations (e.g., 0.84 accuracy on Expert-ECG-QA and 0.76 on SCP-mapped PTB-XL under GPT-4). Overall, CARE-ECG provides traceable reasoning by exposing key latent drivers, causal evidence paths, and how alternative physiological states would change outcomes.
\end{abstract}



\maketitle

\section{Introduction}

\begin{figure*}[t]
    \centering
    \includegraphics[width=\textwidth]{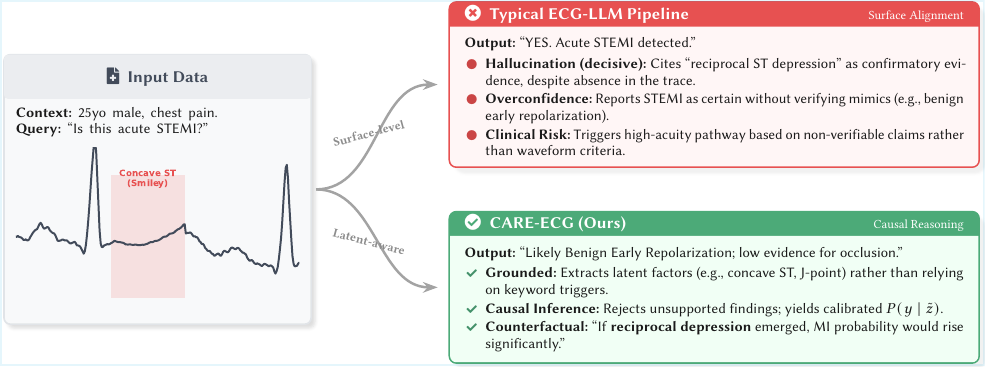}
    \caption{\textbf{Motivation.} Typical ECG-LLMs may over-diagnose STEMI in mimics by hallucinating confirmatory evidence (e.g., reciprocal ST depression). CARE-ECG mitigates this by reasoning over latent morphological biomarkers (e.g., concave ST morphology) and validating decisions via causal counterfactuals.}
    \label{fig:motivation_final}
\end{figure*}

Electrocardiography (ECG) is a cornerstone of cardiovascular diagnosis, providing non-invasive measurements of cardiac electrical activity for detecting arrhythmias, ischemia, and conduction abnormalities~\cite{yu2022ai, murat2020application}. With the rapid digitization of ECG data across hospital monitoring systems, large-scale clinical repositories, and wearable devices~\cite{aqajari2024enhancing, alikhani2024seal, alikhani2024ea}, there is growing interest in automated methods that can assist clinicians in interpreting ECGs accurately and efficiently~\cite{wang2025linkage}. In parallel, large language models (LLMs) have demonstrated strong capabilities in biomedical domains, including clinical report generation, medical question answering, and interactive decision support~\cite{wang2025llm, singhal2025toward, wang2025healthq}. These advances motivate ECG-language systems that convert waveform patterns into narratives for clinicians and support expert queries about diagnostic decisions.

Despite this promise, most existing ECG-LLM systems remain limited in ways that directly affect clinical reliability. Many approaches rely on weakly supervised alignment between ECG signals and text, combined with query-driven retrieval mechanisms that are agnostic to physiological structure and causal dependencies~\cite{yu2023zero, kim2025medical}. While such systems may generate fluent explanations, they often lack robust grounding in clinically relevant evidence and can introduce unsupported claims when faced with expert-level questions~\cite{wang2025healthq}. These weaknesses become evident in evaluation settings that assess both correctness and faithfulness, such as expert ECG question answering and SCP-coded diagnosis grounding, where free-text outputs must align with structured diagnostic evidence rather than surface linguistic similarity~\cite{wang2025medcot}.

Beyond answer quality, a more fundamental limitation concerns reasoning capability. Clinical ECG interpretation is inherently temporal and causal: clinicians reason from waveform morphology to latent physiological states, evaluate how these states evolve over time, and consider counterfactual scenarios such as how QT interval shortening or ST-segment normalization would alter diagnostic risk~\cite{wang2024ecg}. However, current ECG-LLM pipelines provide limited support for such “what-if” analysis and typically do not expose a traceable reasoning structure. As a result, it remains difficult to determine which physiological factors drive predictions, assess sensitivity to signal changes, or systematically verify and mitigate hallucinated explanations in high-stakes clinical settings~\cite{ong2024medical}. Figure~\ref{fig:motivation_final} highlights a representative pitfall: typical ECG-LLMs may over-commit to acute STEMI by hallucinating confirmatory patterns (e.g., reciprocal ST depression) instead of relying on traceable waveform criteria.

These observations motivate ECG-language systems whose reasoning structure more closely mirrors clinical practice and whose outputs can be empirically validated~\cite{wang2025medcot}. Rather than treating ECG interpretation as a direct signal-to-text mapping, an effective system should explicitly represent clinically meaningful factors, preserve temporal organization across cardiac cycles, and provide mechanisms for uncertainty-aware diagnosis~\cite{mckeen2025ecg}. In particular, such systems should encode ECG signals into structured latent biomarkers, reason over them through causal mechanisms that support both probabilistic inference and counterfactual assessment, and generate explanations that remain grounded in retrieved clinical evidence. Importantly, these design principles should translate into measurable improvements under rigorous evaluation---including higher diagnostic accuracy on benchmarked tasks, stronger alignment between generated rationales and causal evidence, and improved robustness to hallucination under expert scrutiny~\cite{wang2025healthq}. Establishing this link between structured reasoning and quantitative gains is essential for advancing ECG-LLMs from fluent assistants to clinically trustworthy decision-support systems~\cite{wang2025ecg}.

To address these challenges, we introduce \textbf{CARE-ECG}, a \textit{Causal Agent-based Reasoning Engine for ECG} that integrates causal representation learning, probabilistic diagnosis, counterfactual reasoning, and grounded language generation in a modular pipeline. CARE-ECG encodes multi-lead ECG waveforms into temporally organized latent biomarkers, enabling reasoning over interpretable physiological factors rather than raw signals alone. It then performs causal graph-based inference to produce diagnostic probabilities and mechanistic rationales, while supporting counterfactual queries that assess how interventions on key biomarkers would change outcomes. To improve explanation faithfulness, CARE-ECG grounds language outputs via retrieval and verification, and uses a modular agentic design to incorporate historical context and maintain traceable reasoning. We evaluate CARE-ECG on PTB-XL~\cite{wagner2020ptb}, MIMIC-IV ECG~\cite{gow2023mimic}, and Expert-ECG-QA~\cite{wang2025ecg} using metrics for answer quality, causal and retrieval alignment, hallucination robustness, and module-level ablations. Results demonstrate consistent improvements over representative ECG-LLM baselines across datasets.

Our contributions are summarized as follows:
\begin{itemize}
    \item We introduce CARE-ECG, a causally structured ECG-language framework that represents ECGs as temporally organized latent biomarkers for interpretable reasoning.
    \item We enable probabilistic diagnosis and counterfactual “what-if” analysis through explicit causal inference over ECG-derived latent factors.
    \item We propose a grounded, modular agentic pipeline that improves explanation faithfulness and reduces hallucinations across ECG benchmarks.
\end{itemize}

\section{Related Work}

\paragraph{LLMs for medical reasoning and grounded clinical assistance.}
LLMs have rapidly advanced medical question answering and decision support, enabling systems that can respond to clinician queries with structured explanations and multi-step reasoning. Notably, Med-PaLM 2~\cite{singhal2025toward} demonstrates that domain-adapted LLMs can achieve expert-level performance on medical QA benchmarks, but also highlights persistent reliability challenges such as hallucinations, incomplete evidence attribution, and overconfident explanations that are difficult to audit in high-stakes settings~\cite{singhal2024medpalm2}. To strengthen reasoning and interpretability, prompting-based scaffolds such as Chain-of-Thought (CoT) have been shown to elicit intermediate reasoning traces that improve multi-step inference and reduce brittle shortcut behavior~\cite{wei2022chain, wang2025medcot}. However, in clinical deployment, fluent reasoning traces are insufficient if they are not \emph{grounded} in verifiable evidence and physiologically plausible mechanisms. This motivates grounding-focused designs—e.g., retrieval augmentation, verification, and structured intermediate states—that constrain generation to clinical criteria and reduce unsupported claims~\cite{shool2025systematic}. CARE-ECG follows this direction by coupling LLM explanations with explicit causal intermediates and evidence-conditioned generation, aiming to improve both reasoning quality and groundedness under expert scrutiny.

\paragraph{ECG interpretation beyond classification: from waveform models to ECG--language systems.}
Deep learning has achieved strong performance in automated ECG interpretation~\cite{wang2024ecg, ebrahimi2020review} using end-to-end waveform models trained for arrhythmia detection and diagnostic classification~\cite{hannun2019cardiologist, ribeiro2020automatic}. These systems show that large-scale supervision can yield highly accurate predictions, yet they typically expose limited mechanisms for \emph{why} a diagnosis is made and how sensitive it is to plausible physiological changes. More recently, multimodal ECG--language modeling has begun bridging ECG signals and natural language to support report-style explanations and interactive query answering, aligning waveform representations with LLM-based generation and dialogue interfaces~\cite{caitowards, mckeen2025ecg}. While promising, many ECG--LLM pipelines still rely on weakly supervised alignment or surface-form retrieval, which can produce clinically fluent but weakly grounded rationales—e.g., hallucinated waveform findings or misapplied thresholds—especially under targeted questioning~\cite{perzhilla2025situ}. CARE-ECG addresses this gap by introducing a structured reasoning layer (latent biomarkers and a causal posterior) that both constrains diagnostic inference and provides an auditable basis for evidence-grounded explanation.

\paragraph{Causal representation learning and counterfactual reasoning for medical explanations.}
Causal modeling offers a principled route toward mechanistic and counterfactual explanations, which are central to clinical reasoning (e.g., assessing how modifying repolarization markers would change risk)~\cite{khatibi2025cdf, asher1976causal}. In representation learning, CausalVAE encourages disentangled latent variables aligned with underlying generative factors, providing a foundation for mapping raw observations to interpretable intermediates that can support downstream causal inference and structured explanations~\cite{yang2021causalvae}. Separately, structured state space models enable efficient modeling of long-range temporal dependencies in sequential data~\cite{gu2021efficiently, dao2024transformers}, matching ECG’s cycle-to-cycle structure and temporal evolution. Building on these ideas, CARE-ECG discretizes latent biomarkers for Bayesian causal graph inference and introduces minimal-edit counterfactuals that expose local diagnostic sensitivity, while grounding language outputs through causal-conditioned retrieval and verifier-triggered fallback. This integration aims to move ECG--LLM systems from post-hoc narrative justification toward clinically aligned reasoning that is both traceable and evidence-linked.

\begin{figure*}[t]
    \centering
    \includegraphics[width=\textwidth]{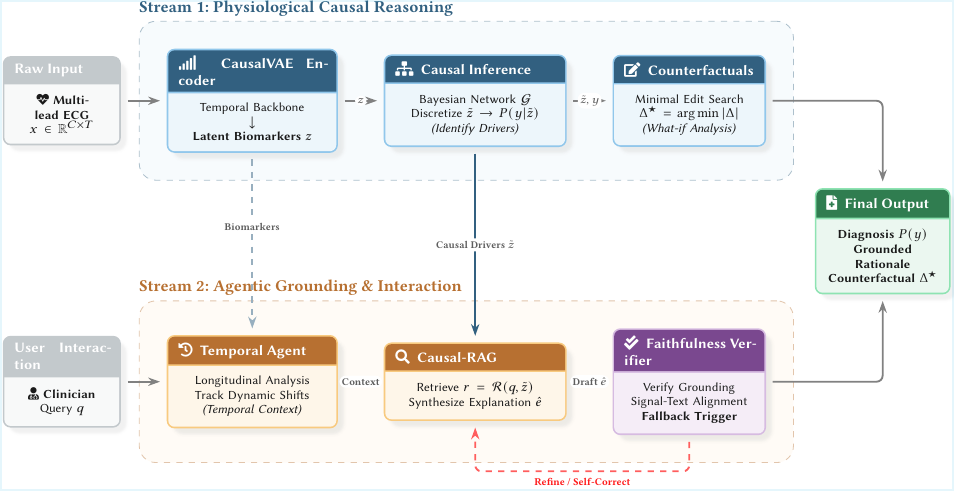}
    \caption{\textbf{The CARE-ECG Framework.} The \textbf{Physiological Stream} encodes ECGs into latent biomarkers and performs causal inference to predict $P(y)$ and enable counterfactual analysis. The \textbf{Agentic Stream} leverages temporal context and causal drivers to guide retrieval and generation, while a \textbf{Faithfulness Verifier} checks signal-grounded alignment and triggers self-correction when needed.}
    \label{fig:care_ecg_framework}
\end{figure*}

\section{Method}

CARE-ECG is a causally structured framework for interpretable ECG-language reasoning that connects signal modeling, causal inference, and language grounding (Fig.~\ref{fig:care_ecg_framework}). The design mirrors clinical practice: clinicians first recognize waveform patterns, then infer latent physiological states, and finally reason about diagnoses and potential interventions. CARE-ECG implements this process through a \textbf{Physiological Stream} (causal representation learning, causal inference, counterfactual reasoning) and an \textbf{Agentic Stream} (causal RAG, faithfulness verification), forming a total of five core components (Fig.~\ref{fig:agentic_reasoning_workflow}).

Formally, we denote a multi-lead ECG segment as $x \in \mathbb{R}^{C \times T}$ and its latent physiological factors as $z \in \mathbb{R}^{d}$; clinical outcomes (diagnostic labels) are denoted by $y$. Given a clinician query $q$, CARE-ECG produces (i) a posterior distribution over diagnoses $P(y \mid x)$, (ii) an evidence-grounded explanation $\hat{e}$, and (iii) optional counterfactual edits over latent factors that clarify how changes in physiology would alter predictions. The overall pipeline can be summarized as
\begin{equation*}
\begin{aligned}
x &\xrightarrow{\text{encoder}} z \xrightarrow{\text{graph}} P(y \mid z), \\
r &= \mathcal{R}(z, y, q), 
\qquad \hat{e} = \mathcal{G}(q, r, z, y),
\end{aligned}
\end{equation*}
where $\mathcal{R}$ retrieves clinical evidence $r$ conditioned on signal-derived factors and graph-inferred hypotheses, and $\mathcal{G}$ generates explanations grounded in retrieved documents and causal evidence. In the following, we describe each component and its role in producing faithful and traceable ECG-language outputs.

\subsection{Causal ECG Representation Learning}

The ECG encoder transforms raw waveforms into a structured latent space intended to capture clinically meaningful factors. Given an input signal $x \in \mathbb{R}^{C \times T}$ (12-lead, 10s recordings sampled at 500\,Hz), we learn a latent representation $z \in \mathbb{R}^d$ using a CausalVAE~\cite{yang2020causalvae}. The key motivation is to move beyond black-box embeddings toward disentangled latent factors that can serve as interpretable intermediates for downstream causal inference and counterfactual analysis. Concretely, the encoder first summarizes waveform dynamics into a fixed-length representation $h$ through a temporal backbone $f_{\text{temp}}(\cdot)$, followed by mean and variance heads that parameterize a Gaussian posterior:
\begin{equation*}
    h = f_{\text{temp}}(x), \quad \mu = W_{\mu} h, \quad \log \sigma^2 = W_{\sigma} h, \quad z = \mu + \sigma \odot \epsilon, \ \epsilon \sim \mathcal{N}(0, I).
\end{equation*}
In our implementation, $f_{\text{temp}}$ can be instantiated with Mamba-style sequence modeling or a BiLSTM fallback, providing a lightweight but expressive temporal summary across cardiac cycles. The decoder reconstructs the waveform from $z$, enabling end-to-end representation learning. We optimize the standard evidence lower bound:
\begin{equation*}
    \mathcal{L}_{\text{CVAE}} = \mathbb{E}_{q_{\phi}(z|x)} \left[ \log p_{\theta}(x|z) \right] - D_{\mathrm{KL}}(q_{\phi}(z|x) \,\|\, p(z)),
\end{equation*}
where $q_{\phi}(z|x)$ and $p_{\theta}(x|z)$ are the encoder and decoder distributions, and $p(z) = \prod_i p(z_i)$ uses a factorized prior to encourage independence among latent factors. This construction promotes compactness and interpretability while retaining high-fidelity reconstruction, and it enables downstream reasoning by exposing a structured representation derived from waveform dynamics.

To interface with probabilistic causal graphs, we discretize each latent dimension into $K$ quantile bins, denoted $\tilde{z}_i \in \{1,\dots,K\}$. This yields a discrete representation that supports efficient Bayesian network learning and inference while preserving ordinal information (e.g., low vs.\ high biomarker regimes). The discretized factors $\tilde{z}$ serve as the primary evidence variables for causal graph inference and counterfactual edits.

\subsection{Bayesian Causal Graph Inference}

To reason about latent biomarkers and disease outcomes, CARE-ECG constructs a causal Bayesian network $\mathcal{G} = (\mathcal{V}, \mathcal{E})$, where $\mathcal{V}$ includes discretized latent factors $\tilde{z}$ and clinical labels $y$, and $\mathcal{E}$ encodes directed dependencies. The model factorizes the joint distribution as
\begin{equation*}
    P(\mathbf{y}, \tilde{\mathbf{z}}) = \prod_{v \in \mathcal{V}} P(v \mid \mathrm{Pa}(v)),
\end{equation*}
where $\mathrm{Pa}(v)$ denotes the parents of node $v$ in the graph. The causal graph provides two practical advantages: (i) it yields posterior diagnoses via structured probabilistic inference rather than opaque classification, and (ii) it exposes evidence paths that can be mapped to explanations.

We initialize the graph structure using clinical priors (e.g., known relationships between rhythm irregularity, interval prolongation, and arrhythmia likelihood) and refine it using score-based structure search (e.g., K2) over the discretized latent variables. Conditional probability tables are learned with Bayesian estimation, which improves stability under limited supervision and supports calibrated posteriors. Given observed evidence $\tilde{z}$, CARE-ECG infers posterior disease probabilities $P(y \mid \tilde{z})$ via variable elimination:
\begin{equation*}
    P(y \mid \tilde{z}) \propto \sum_{\mathcal{V} \setminus \{y\}} \prod_{v \in \mathcal{V}} P(v \mid \mathrm{Pa}(v)).
\end{equation*}
This inference step produces both a diagnosis distribution and a ranked set of contributing latent factors, enabling traceable evidence chains that later guide retrieval and explanation generation.

\subsection{Counterfactual Reasoning Engine}

CARE-ECG supports clinically meaningful counterfactual questions such as: ``If the QT interval were shortened, would MI risk decrease?'' We operationalize counterfactual reasoning over the learned Bayesian graph by searching for minimal evidence edits that achieve a desired diagnostic outcome. The goal is to provide actionable, interpretable explanations in the form of small changes to latent physiological factors that alter the posterior prediction.

Given observed evidence $e$ (typically the discretized latent factors $\tilde{z}$) and a desired target value $y^*$, we solve
\begin{equation*}
    \Delta^\star = \arg\min_{\Delta} \ |\Delta|
    \ \ \text{s.t.} \ \ 
    \arg\max_y P(y \mid e \oplus \Delta) = y^*,
\end{equation*}
where $\Delta$ denotes a set of discrete edits to evidence variables and $|\Delta|$ is the number of edited factors. Here, $e \oplus \Delta$ indicates replacing selected evidence variables with alternative discrete states (e.g., moving a biomarker from ``High'' to ``Mid''). In practice, CARE-ECG performs a single-variable search over candidate factors and returns the minimal edit when possible, along with the updated posterior distribution. This produces counterfactuals that remain local and clinically interpretable: they communicate how a small physiological shift could plausibly change diagnostic conclusions, while avoiding unrealistic multi-variable interventions.

\subsection{Causal-RAG: Knowledge-Grounded Inference}

To enhance factual correctness and reduce hallucinations, CARE-ECG incorporates a causal retrieval-augmented generation module~\cite{khatibi2025cdf}. Unlike standard RAG that retrieves based solely on query semantics, CARE-ECG conditions retrieval on signal-derived factors and graph-inferred hypotheses. We maintain an offline knowledge index of ECG-relevant documents (e.g., guideline statements, textbook excerpts, SCP mappings) embedded using a sentence-transformer model. At inference time, the retrieval query is enriched with clinically meaningful terms derived from top contributing latent factors and posterior diagnoses:
\begin{equation*}
    r = \mathcal{R}(\text{enrich}(q, \tilde{z}, y)), 
    \qquad 
    \hat{e} = \mathcal{G}(q, r, \tilde{z}, y).
\end{equation*}
This design encourages the generator to cite and reflect causally relevant evidence (e.g., interval thresholds, waveform criteria, guideline patterns) rather than generating explanations from surface correlations alone.

To further improve reliability, we compute a hallucination risk score based on fuzzy matching between retrieved facts and the generated explanation:
\begin{equation*}
    \mathrm{HR} = 1 - \frac{1}{|r|} \sum_{f \in r} \mathbf{1}\{\mathrm{match}(f, \hat{e})\}.
\end{equation*}
If $\mathrm{HR}$ exceeds a threshold, CARE-ECG falls back to a retrieval-only explanation mode that prioritizes factual grounding, returning a concise evidence-based rationale. This mechanism aligns with our evaluation protocol for hallucination robustness and provides a practical safety layer for clinical-facing generation.

\subsection{Agentic Reasoning and Verification}

\begin{figure}[t]
    \centering
    \includegraphics[width=\linewidth]{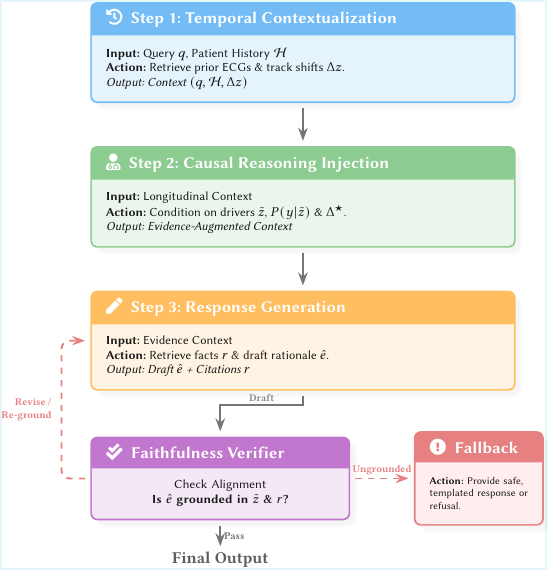}
    \caption{\textbf{Agentic reasoning workflow.} CARE-ECG integrates temporal context ($\Delta z$) and causal evidence ($\tilde{z}$) before generating a grounded rationale, verified by a faithfulness gate with self-correction and fallback.}
    \label{fig:agentic_reasoning_workflow}
\end{figure}


\definecolor{cTabHeader}{HTML}{E8F4F8}  
\definecolor{cTabGroup}{HTML}{F7F9FA}   
\definecolor{cVar}{HTML}{D81B60}        
\definecolor{cCode}{HTML}{2C3E50}       

\begin{table}[t]
\centering
\footnotesize 
\renewcommand{\arraystretch}{1.3} 
\setlength{\tabcolsep}{3pt}       

\begin{tabularx}{\linewidth}{@{}p{0.25\linewidth}X@{}}
\toprule
\rowcolor{cTabHeader}
\textbf{Module / Field} & \textbf{Prompted Instruction \& Runtime Logic} \\ 
\midrule

\multicolumn{2}{@{}l}{\cellcolor{cTabGroup}\textit{\textbf{1. Prompt Construction}}} \\ 
\addlinespace[2pt]

Patient Query & 
Injects: \texttt{Patient Query: \textcolor{cVar}{\{query\}}} \\

Causal Factors & 
Injects: \texttt{Key Causal Factors (from VAE/Graph): \textcolor{cVar}{\{formatted\_causal\}}}. \newline
\textit{(Logic: Comma-joined list of drivers; returns \texttt{'None'} if empty).} \\

Retrieved Evidence & 
Injects: \texttt{Retrieved Medical Facts (RAG):} followed by enumerated list: \texttt{[Fact i] \textcolor{cVar}{\{fact\_i\}}}. \\

Gen. Instruction & 
Static string: ``Explain the prediction \textit{clearly and medically grounded}, and attach citations using fact tags (e.g., \texttt{[Fact 1]}).'' \\

\midrule
\multicolumn{2}{@{}l}{\cellcolor{cTabGroup}\textit{\textbf{2. Generation Parameters}}} \\ 
\addlinespace[2pt]

LLM Config & 
Call \texttt{openai.chat.completions} with \texttt{model="gpt-4"}, \texttt{temperature=0.3}, \texttt{max\_tokens=600}. \newline
Output: Draft explanation $\hat{e}$. \\

\midrule
\multicolumn{2}{@{}l}{\cellcolor{cTabGroup}\textit{\textbf{3. Safety \& Verification Logic}}} \\ 
\addlinespace[2pt]

Scoring Function & 
Compute \texttt{halluc\_score} via fuzzy logic matching: \newline
\texttt{score\_hallucination\_risk\_fuzzy(explanation, rag\_facts)}. \\

Safety Gate & 
\textbf{Trigger Fallback IF:} \newline
\texttt{halluc\_score > 0.5} \textbf{AND} \texttt{fallback\_to\_rag\_only=True}. \\

Fallback Action & 
1. Re-prompt with \texttt{causal\_facts=[]} (Force RAG-only). \newline
2. Append warning: \texttt{"(Note: Fallback used due to high hallucination risk.)"} \\

Return Payload & 
Returns dict: \texttt{\{explanation, hallucination\_score, used\_fallback\}}. \newline
\textit{(Logs \texttt{raw\_with\_causal} and \texttt{raw\_rag\_only} for audit)}. \\

\bottomrule
\end{tabularx}
\caption{\textbf{Response Agent prompt \& safety gate.} The table summarizes the fact-tagged prompt template, decoding settings, hallucination scoring, and the RAG-only fallback rule used for grounded explanation generation.}
\label{tab:care_ecg_prompt_logic}
\end{table}

Within CARE-ECG, we instantiate the agentic reasoning component as a modular pipeline that exposes intermediate states for verification (Fig.~\ref{fig:agentic_reasoning_workflow}). 
Instead of using a single end-to-end generation prompt, the system decomposes reasoning into three agents with explicit responsibilities: 
a History Agent that retrieves patient-specific prior ECGs when available (e.g., in MIMIC-IV ECG) or semantically similar ECGs as surrogate history when longitudinal records are unavailable, and summarizes longitudinal shifts in latent factors $\Delta z$; 
a Diagnosis Agent that integrates causal graph inference with counterfactual checks to form a structured diagnostic state; 
and a Response Agent that generates grounded explanations conditioned on retrieved evidence.
The agents communicate via a compact message tuple,
\begin{equation*}
    m = \langle q, \Delta z, \tilde{z}, \hat{y}, r \rangle ,
\end{equation*}
which contains the clinician query $q$, longitudinal change $\Delta z$, discretized biomarkers $\tilde{z}$, the inferred diagnostic prediction $\hat{y}$, and retrieved facts $r$.
This tuple is used to assemble a fact-tagged prompt (Tab.~\ref{tab:care_ecg_prompt_logic}), encouraging the Response Agent to explicitly attribute clinical claims to retrieved evidence while retaining the salient causal drivers highlighted by the Diagnosis Agent.

To improve faithfulness under expert-level questioning, the Response Agent enforces an explicit verification gate. 
Given a draft explanation $\hat{e}$, we compute a hallucination risk score $\mathrm{HR}(\hat{e}, r)$ via fuzzy matching against retrieved facts and trigger a retrieval-only regeneration when the score exceeds a predefined threshold. 
This yields a lightweight self-correction loop,
$m \rightarrow \hat{e} \rightarrow \mathrm{HR}(\hat{e}, r)$,
that is consistent with the runtime behavior in Fig.~\ref{fig:agentic_reasoning_workflow} and the prompt specification in Tab.~\ref{tab:care_ecg_prompt_logic}.

\subsection{Output Specification and Use Modes}

The final output includes disease probability distributions, rationale summaries, counterfactual edits, and optional multi-turn dialogues for interactive clinical use. Outputs are returned in structured formats (e.g., JSON) with optional natural language for clinical review and EHR integration. Intermediate representations (latent factors, causal evidence paths, retrieved facts, hallucination scores, and counterfactual edits) are retained for auditability and retrospective analysis.

\section{Experiments and Results}


\definecolor{cHead}{HTML}{E8F4F8}  
\definecolor{cGroup}{HTML}{F7F9FA} 

\begin{table*}[t]
\centering
\caption{\textbf{Evaluation datasets and protocols.} 
PTB-XL evaluates diagnosis grounding via Text$\rightarrow$SCP mapping, 
MIMIC-IV ECG evaluates retrieval-based faithfulness, 
and Expert-ECG-QA evaluates interactive clinical reasoning.}
\label{tab:dataset_summary_final_v2}
\small
\renewcommand{\arraystretch}{1.25}
\setlength{\tabcolsep}{3pt}

\begin{tabularx}{0.98\textwidth}{@{} l >{\raggedright\arraybackslash}X >{\raggedright\arraybackslash}X >{\raggedright\arraybackslash}X @{}}
\toprule
\rowcolor{cHead}
\textbf{Features} & \textbf{PTB-XL} & \textbf{MIMIC-IV ECG} & \textbf{Expert-ECG-QA} \\
\midrule

\rowcolor{cGroup}
\multicolumn{4}{l}{\textit{\textbf{Data Characteristics}}} \\

\textbf{Content} & 
Standard 12-lead ECGs with SCP-ECG statements. & 
Large-scale clinical ECG corpus for retrieval testing. & 
Multimodal QA pairs (Real + Synthetic) covering 12 tasks. \\

\textbf{Size} & 
\textbf{21,837} records \newline (18,885 patients) & 
\textbf{$\approx$800,000} records \newline ($\approx$160,000 patients) & 
\textbf{47,211} QA pairs \newline (Multi-turn supported) \\

\textbf{Input $\rightarrow$ GT} & 
Signal $x \rightarrow$ SCP Codes & 
Signal $x \rightarrow$ Clinical Text & 
Signal $x$ + Query $q \rightarrow$ Answer \\

\midrule

\rowcolor{cGroup}
\multicolumn{4}{l}{\textit{\textbf{Evaluation Protocol}}} \\

\textbf{Core Task} & 
\textbf{Diagnosis Grounding} \newline (Text $\rightarrow$ Code Mapping) & 
\textbf{Faithfulness Check} \newline (Retrieval Alignment) & 
\textbf{Clinical Reasoning} \newline (Interactive QA) \\

\textbf{Metrics} & 
Macro-F1, Coverage, CRC & 
Groundedness, Hallucination Rate & 
Exact Match, SRS, CRC \\

\textbf{Reliability} & 
Structured Correctness & 
Robustness to Noise & 
Expert-level Logic \\

\bottomrule
\end{tabularx}
\end{table*}

Our experiments are designed to evaluate CARE-ECG as a \emph{clinical reasoning system} rather than a purely generative model. In ECG interpretation, correctness alone is insufficient: an answer may be diagnostically correct yet justified with fabricated waveform findings (e.g., hallucinated ST elevation), incorrect thresholds (e.g., QTc cutoffs), or irrelevant guideline statements. Such failures are especially harmful because they present as confident mechanistic explanations. We therefore structure the evaluation around three reliability questions: \textbf{(Q1) clinical correctness} under expert QA and structured diagnosis grounding, \textbf{(Q2) faithfulness} to causal drivers and retrieved evidence, and \textbf{(Q3) robustness} against hallucinations. This organization mirrors how clinicians assess decision-support tools: Does it get the answer right? Does it cite the right physiological reasons? And does it avoid inventing evidence under pressure?

Across all datasets, we keep inference-time conditions controlled to ensure comparisons reflect reasoning and grounding design rather than decoding artifacts. All models use the same context length and decoding hyperparameters, and baselines are configured with comparable retrieval budgets where retrieval is applicable. We evaluate with two backbones (GPT-4 and LLaMA 3--8B) to probe whether the gains arise from CARE-ECG’s structured pipeline rather than from any single LLM’s built-in medical knowledge. Importantly, this also reflects practical deployment: stronger proprietary LLMs may be available in some settings, while open models are preferred in others due to privacy and cost constraints.

\subsection{Experimental Setup}
We evaluate on three datasets that emphasize complementary clinical behaviors (Table~\ref{tab:dataset_summary_final_v2}). 
\textbf{PTB-XL} supports structured diagnosis grounding via SCP-ECG statements: the model produces free-text interpretations that are mapped to SCP codes using keyword extraction and fuzzy matching. 
This protocol is intentionally strict---it rewards outputs that commit to clinically specific diagnostic statements and penalizes generic narrative completion. 
\textbf{MIMIC-IV ECG} serves as a large-scale clinical corpus where we evaluate groundedness and hallucination robustness under retrieval-based settings, which stress whether generated rationales remain consistent with retrieved clinical evidence when explicit QA-style supervision is limited. 
\textbf{Expert-ECG-QA} further provides expert-validated multimodal question answering, including multi-turn interactions, and is used to assess whether a system can sustain faithful clinical reasoning under targeted questioning about waveform evidence and counterfactual changes.

CARE-ECG integrates five components: (i) latent biomarker inference (CausalVAE + temporal backbone), (ii) Bayesian causal graph inference over discretized factors $\tilde{z}$, (iii) counterfactual minimal edits $\Delta^\star$ for “what-if” assessment, (iv) causal-conditioned retrieval that enriches queries with dominant causal factors and inferred hypotheses, and (v) verifier-triggered fallback that suppresses hallucination when evidence alignment is weak. In contrast, baselines represent typical ECG--LLM pipelines that emphasize surface alignment and/or shallow retrieval without explicit causal structuring. Our goal is not to claim that any baseline is “bad,” but to measure whether explicit causal structure plus evidence control yields measurable gains on correctness, faithfulness, and robustness.

\subsection{Evaluation Metrics}
We report four groups of metrics, each targeting a distinct clinical failure mode. \textbf{Answer Quality} is measured by Accuracy, Precision, Recall, and F1 on datasets with ground truth labels: Expert-ECG-QA directly, and PTB-XL after SCP mapping. These metrics measure whether the system arrives at correct diagnostic content. For MIMIC-IV ECG, we omit these metrics because the dataset lacks explicit QA-style annotations or structured diagnosis labels that can be directly aligned with free-text LLM outputs. More importantly, answer correctness alone is insufficient for clinical deployment. A system may produce correct conclusions for incorrect reasons, or hallucinate plausible but unsupported waveform findings—failure modes that pose significant clinical risk. We therefore introduce two additional groups of metrics to evaluate grounding and hallucination behavior.


\textbf{Causal and Retrieval Alignment} is measured by Causal Retrieval Coverage (CRC), Groundedness, and Context Relevance. CRC measures whether the explanation explicitly reflects the top causal drivers implied by the latent causal graph; this mitigates the risk of post-hoc rationalization where the model “talks around” the answer without referencing the physiological factors that actually supported the inference. Groundedness measures alignment between the generated explanation and retrieved documents, and Context Relevance measures semantic alignment to the clinician query, discouraging generic guideline dumping. Together, these metrics quantify whether the explanation is both \emph{causally anchored} and \emph{evidence-linked}.


\begin{table*}[t]
\centering
\caption{Answer Quality Metrics on Expert-ECG-QA and PTB-XL. For PTB-XL, free-text outputs were mapped to SCP diagnostic codes using keyword extraction and fuzzy label matching. \textbf{Bold} indicates best performance.}
\label{tab:answer_quality}
\small
\renewcommand{\arraystretch}{1.2} 
\setlength{\tabcolsep}{6pt} 

\begin{tabular}{l cccc cccc}
\toprule
\multirow{2.5}{*}{\textbf{Model}} & \multicolumn{4}{c}{\textbf{GPT-4 Backbone}} & \multicolumn{4}{c}{\textbf{LLaMA 3--8B Backbone}} \\
\cmidrule(lr){2-5} \cmidrule(lr){6-9}
 & Acc & Prec & Rec & F1 & Acc & Prec & Rec & F1 \\
\midrule

\multicolumn{9}{l}{\textit{\textbf{Dataset: Expert-ECG-QA}}} \\
\rowcolor{cHighlight}
\textbf{CARE-ECG (Ours)} & \textbf{0.84} & \textbf{0.79} & \textbf{0.87} & \textbf{0.83} & \textbf{0.77} & \textbf{0.73} & \textbf{0.80} & \textbf{0.76} \\
ECG-Chat & 0.76 & 0.72 & 0.80 & 0.75 & 0.71 & 0.68 & 0.73 & 0.70 \\
Q-HEART & 0.74 & 0.70 & 0.77 & 0.73 & 0.69 & 0.66 & 0.70 & 0.68 \\
anyECG-Chat & 0.71 & 0.68 & 0.72 & 0.70 & 0.66 & 0.64 & 0.67 & 0.65 \\
\addlinespace[4pt]

\multicolumn{9}{l}{\textit{\textbf{Dataset: PTB-XL (SCP-Mapped)}}} \\
\rowcolor{cHighlight}
\textbf{CARE-ECG (Ours)} & \textbf{0.76} & \textbf{0.73} & \textbf{0.78} & \textbf{0.75} & \textbf{0.70} & \textbf{0.68} & \textbf{0.72} & \textbf{0.70} \\
ECG-Chat & 0.70 & 0.68 & 0.73 & 0.70 & 0.66 & 0.63 & 0.69 & 0.66 \\
Q-HEART & 0.68 & 0.66 & 0.70 & 0.68 & 0.64 & 0.61 & 0.65 & 0.63 \\
anyECG-Chat & 0.66 & 0.63 & 0.68 & 0.65 & 0.62 & 0.60 & 0.64 & 0.61 \\

\bottomrule
\end{tabular}
\end{table*}

\textbf{Hallucination and Factuality} are measured by Hallucination Rate (HR; lower is better) and Semantic Relevance Score (SRS; higher is better). HR penalizes unsupported claims relative to retrieved evidence, capturing whether the system invents waveform findings, thresholds, or guideline statements. SRS complements this by evaluating whether the explanation remains semantically consistent with evidence rather than merely shorter or less specific. Finally, we report an \textbf{ablation study} to isolate the contribution of each CARE-ECG module while keeping all other factors fixed.

\subsection{Results: Clinical correctness under expert QA and structured grounding (Q1)}
Table~\ref{tab:answer_quality} summarizes answer quality on Expert-ECG-QA and PTB-XL (SCP-mapped). CARE-ECG achieves the best performance under both GPT-4 and LLaMA 3--8B across both datasets. On Expert-ECG-QA, CARE-ECG reaches \textbf{0.84 accuracy} under GPT-4, outperforming ECG-Chat (0.76), Q-HEART (0.74), and anyECG-Chat (0.71). The same trend holds for precision/recall/F1, indicating that gains are not driven by a particular operating point (e.g., overly conservative answers): the system improves both hit rate and overall balance. Under LLaMA 3--8B, CARE-ECG similarly leads with \textbf{0.77 accuracy}, again exceeding all baselines.

On PTB-XL, where evaluation requires mapping the generated interpretation to SCP codes, CARE-ECG also leads (GPT-4 accuracy \textbf{0.76}; LLaMA accuracy \textbf{0.70}). This setting is informative because it is less forgiving than free-form similarity metrics: the system must generate diagnostic statements that survive structured mapping. The consistent improvements across both Expert-ECG-QA and PTB-XL suggest that CARE-ECG’s causal representation and graph-based inference help produce outputs that are not only plausible narratives but also structurally consistent with diagnostic coding schemes.

A key takeaway from Table~\ref{tab:answer_quality} is that improvements persist across backbones. If the gains were solely due to stronger language priors, we would expect a larger gap under GPT-4 and a collapse under LLaMA. Instead, CARE-ECG maintains leading performance under LLaMA 3--8B, indicating that the structured pipeline contributes meaningfully by constraining the generation process around latent factors, causal inference outputs, and retrieved evidence.

\begin{table*}[t]
\centering
\caption{Causal and Retrieval Metrics on PTB-XL, MIMIC-IV ECG, and Expert-ECG-QA. \textbf{Bold} denotes the best performance.}
\label{tab:causal_retrieval}
\small 
\renewcommand{\arraystretch}{1.2} 
\setlength{\tabcolsep}{5pt} 

\begin{tabular}{l ccc ccc}
\toprule
\multirow{2.5}{*}{\textbf{Model}} & \multicolumn{3}{c}{\textbf{GPT-4 Backbone}} & \multicolumn{3}{c}{\textbf{LLaMA 3--8B Backbone}} \\
\cmidrule(lr){2-4} \cmidrule(lr){5-7}
 & CRC & Ground. & Cont.Rel. & CRC & Ground. & Cont.Rel. \\
\midrule

\multicolumn{7}{l}{\textit{\textbf{Dataset: PTB-XL}}} \\
\rowcolor{cHighlight}
\textbf{CARE-ECG (Ours)} & \textbf{0.91} & \textbf{0.71} & \textbf{0.75} & \textbf{0.86} & \textbf{0.67} & \textbf{0.71} \\
ECG-Chat & 0.78 & 0.62 & 0.64 & 0.73 & 0.56 & 0.59 \\
Q-HEART & 0.75 & 0.60 & 0.61 & 0.71 & 0.54 & 0.57 \\
anyECG-Chat & 0.72 & 0.57 & 0.58 & 0.69 & 0.52 & 0.55 \\
\addlinespace[4pt] 

\multicolumn{7}{l}{\textit{\textbf{Dataset: MIMIC-IV ECG}}} \\
\rowcolor{cHighlight}
\textbf{CARE-ECG (Ours)} & \textbf{0.94} & \textbf{0.73} & \textbf{0.76} & \textbf{0.87} & \textbf{0.69} & \textbf{0.72} \\
ECG-Chat & 0.81 & 0.64 & 0.67 & 0.76 & 0.59 & 0.62 \\
Q-HEART & 0.78 & 0.61 & 0.63 & 0.72 & 0.57 & 0.59 \\
anyECG-Chat & 0.76 & 0.60 & 0.61 & 0.70 & 0.55 & 0.58 \\
\addlinespace[4pt]

\multicolumn{7}{l}{\textit{\textbf{Dataset: Expert-ECG-QA}}} \\
\rowcolor{cHighlight}
\textbf{CARE-ECG (Ours)} & \textbf{0.96} & \textbf{0.75} & \textbf{0.78} & \textbf{0.90} & \textbf{0.71} & \textbf{0.74} \\
ECG-Chat & 0.83 & 0.65 & 0.68 & 0.77 & 0.61 & 0.64 \\
Q-HEART & 0.80 & 0.63 & 0.65 & 0.73 & 0.58 & 0.60 \\
anyECG-Chat & 0.78 & 0.60 & 0.63 & 0.70 & 0.54 & 0.57 \\

\bottomrule
\end{tabular}
\end{table*}

\subsection{Results: Faithfulness to causal drivers and retrieved evidence (Q2)}
Clinical deployment requires explanations that can be audited: clinicians must be able to trace \emph{why} a diagnosis was suggested and whether that rationale matches evidence. Table~\ref{tab:causal_retrieval} evaluates this property using CRC (causal alignment), groundedness (evidence overlap), and context relevance (query alignment). CARE-ECG achieves the highest scores across all three datasets and both backbones.

Under GPT-4, CARE-ECG achieves CRC of \textbf{0.91} on PTB-XL, \textbf{0.94} on MIMIC-IV ECG, and \textbf{0.96} on Expert-ECG-QA, exceeding ECG-Chat (0.78/0.81/0.83) and other baselines. This indicates that explanations more frequently mention and operationalize the causal factors identified by the model’s own inference, reducing the chance of “decorative” explanations that sound clinical but are decoupled from the model’s reasoning. Groundedness follows the same pattern (CARE-ECG: 0.71/0.73/0.75), supporting the claim that retrieval is not merely appended but actively used to shape the explanation content.

\begin{figure*}[t]
\centering
\includegraphics[width=\textwidth]{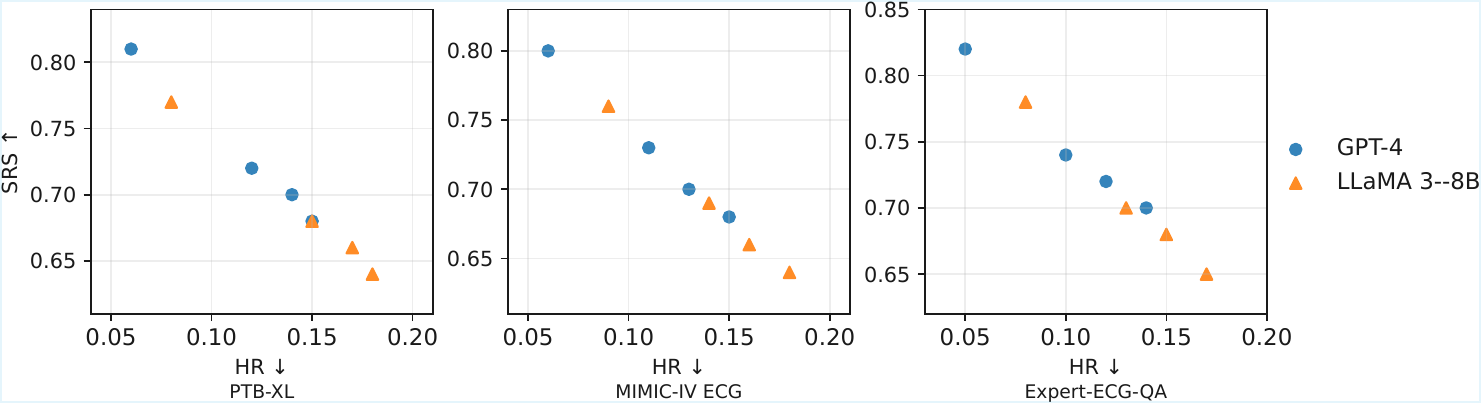}
\caption{\textbf{Faithfulness--robustness trade-off in the HR--SRS plane across datasets.} Each panel shows HR (lower is better) versus SRS (higher is better) for all methods under two backbones. Points closer to the upper-left indicate more evidence-consistent explanations with fewer hallucinations.}
\label{fig:hr_srs_scatter}
\end{figure*}

Context relevance is also consistently higher for CARE-ECG (GPT-4: 0.75/0.76/0.78; LLaMA: 0.71/0.72/0.74). This is important because adding retrieval can sometimes hurt query focus by encouraging generic medical text. The fact that context relevance improves suggests that causal conditioning helps retrieval remain query-specific: the system does not simply retrieve “common ECG facts,” but retrieves evidence linked to the current latent regime and inferred hypothesis.

Overall, Table~\ref{tab:causal_retrieval} suggests that the benefit of CARE-ECG is not merely improved retrieval coverage, but a stronger \emph{constraint structure} on explanation formation. 
Across all three benchmarks, CARE-ECG achieves consistently higher CRC, groundedness, and context relevance than prior ECG-LLM baselines under both GPT-4 and LLaMA~3--8B backbones, indicating that its rationales are more aligned with causal drivers and retrieved evidence. 
Notably, the gains are simultaneous across the three metrics, rather than trading off one criterion for another (e.g., retrieving more text but reducing relevance), which supports the interpretation that causal inference and retrieval play complementary roles: causal reasoning prioritizes which latent drivers should matter, retrieval supplies the external supporting facts, and the response generation stage remains anchored to both signals.

From a clinical perspective, this pattern is stronger than “better explanations” in a subjective sense: it indicates that CARE-ECG produces rationales that are \emph{more checkable} against explicit evidence constraints, instead of relying on narrative plausibility. 
The same trend holds in Expert-ECG-QA, where answers must remain consistent under targeted questioning, as well as in PTB-XL where outputs are evaluated against structured diagnostic evidence, suggesting that the proposed causal-and-verification design improves reliability across both grounding-oriented and interactive reasoning settings.

\subsection{Results: Hallucination robustness and factuality (Q3)}

Beyond correctness and alignment, clinical safety hinges on whether the system fabricates evidence, especially under expert questioning. Table~\ref{tab:hallucination_factuality} reports HR (lower is better) and SRS (higher is better). CARE-ECG achieves the lowest hallucination rate across all datasets and both backbones. Under GPT-4, CARE-ECG obtains HR of \textbf{0.06} on PTB-XL, \textbf{0.06} on MIMIC-IV ECG, and \textbf{0.05} on Expert-ECG-QA, improving substantially over ECG-Chat (0.12/0.11/0.10) and other baselines. Under LLaMA 3--8B, CARE-ECG likewise achieves lower HR (0.08/0.09/0.08) than baselines, suggesting that hallucination suppression is not purely a property of the backbone LLM.

Crucially, CARE-ECG’s lower HR does not come at the cost of vagueness. SRS remains highest for CARE-ECG under GPT-4 (0.81/0.80/0.82) and under LLaMA (0.77/0.76/0.78). This indicates that the system maintains semantically rich explanations while better aligning them with evidence, rather than simply producing shorter or more hedged outputs. In practical terms, this behavior matches the intended role of the verifier-triggered fallback: when evidence alignment is weak, the system is pushed toward evidence-first generation rather than free-form completion.


\begin{table*}[t]
\centering
\caption{Hallucination and Factuality Metrics. \textbf{HR}: Hallucination Rate (lower $\downarrow$ is better); \textbf{SRS}: Summary Relevance Score (higher $\uparrow$ is better). \textbf{Bold} indicates best performance.}
\label{tab:hallucination_factuality}
\small
\renewcommand{\arraystretch}{1.2} 
\setlength{\tabcolsep}{10pt} 

\begin{tabular}{l cc cc}
\toprule
\multirow{2.5}{*}{\textbf{Model}} & \multicolumn{2}{c}{\textbf{GPT-4 Backbone}} & \multicolumn{2}{c}{\textbf{LLaMA 3--8B Backbone}} \\
\cmidrule(lr){2-3} \cmidrule(lr){4-5}
 & HR ($\downarrow$) & SRS ($\uparrow$) & HR ($\downarrow$) & SRS ($\uparrow$) \\
\midrule

\multicolumn{5}{l}{\textit{\textbf{Dataset: PTB-XL}}} \\
\rowcolor{cHighlight}
\textbf{CARE-ECG (Ours)} & \textbf{0.06} & \textbf{0.81} & \textbf{0.08} & \textbf{0.77} \\
ECG-Chat & 0.12 & 0.72 & 0.15 & 0.68 \\
Q-HEART & 0.14 & 0.70 & 0.17 & 0.66 \\
anyECG-Chat & 0.15 & 0.68 & 0.18 & 0.64 \\
\addlinespace[4pt]

\multicolumn{5}{l}{\textit{\textbf{Dataset: MIMIC-IV ECG}}} \\
\rowcolor{cHighlight}
\textbf{CARE-ECG (Ours)} & \textbf{0.06} & \textbf{0.80} & \textbf{0.09} & \textbf{0.76} \\
ECG-Chat & 0.11 & 0.73 & 0.14 & 0.69 \\
Q-HEART & 0.13 & 0.70 & 0.16 & 0.66 \\
anyECG-Chat & 0.15 & 0.68 & 0.18 & 0.64 \\
\addlinespace[4pt]

\multicolumn{5}{l}{\textit{\textbf{Dataset: Expert-ECG-QA}}} \\
\rowcolor{cHighlight}
\textbf{CARE-ECG (Ours)} & \textbf{0.05} & \textbf{0.82} & \textbf{0.08} & \textbf{0.78} \\
ECG-Chat & 0.10 & 0.74 & 0.13 & 0.70 \\
Q-HEART & 0.12 & 0.72 & 0.15 & 0.68 \\
anyECG-Chat & 0.14 & 0.70 & 0.17 & 0.65 \\

\bottomrule
\end{tabular}
\end{table*}

The combined pattern in Table~\ref{tab:hallucination_factuality} suggests that CARE-ECG improves clinical reliability in two coupled ways: it reduces unsupported claims (HR) and increases semantic consistency with evidence (SRS). This is especially important in ECG settings where plausible-sounding but incorrect waveform descriptions can be difficult for non-experts to detect.

To complement Table~\ref{tab:hallucination_factuality} with a more geometric view, Fig.~\ref{fig:hr_srs_scatter} visualizes the joint trade-off between hallucination rate (HR) and semantic relevance score (SRS). Unlike reporting each metric independently, the scatter perspective makes it clear whether a method achieves low hallucination by becoming overly conservative (which would typically reduce SRS) or whether it can simultaneously suppress unsupported claims while maintaining semantically rich, evidence-aligned explanations. Across all datasets and both backbones, CARE-ECG occupies the favorable region characterized by \emph{lower HR and higher SRS}, consistent with the numerical dominance shown in Table~\ref{tab:hallucination_factuality}. This indicates that the gains are not driven by truncation or hedging, but by shifting generation toward evidence-consistent content through causal conditioning and verifier-triggered fallback.

\begin{table*}[t]
\centering
\caption{Ablation Study on Expert-ECG-QA (GPT-4 vs. LLaMA 3--8B). ``Latent'' denotes CausalVAE+temporal backbone; ``Graph'' denotes Bayesian inference over $\tilde{z}$; ``RAG'' denotes causal-conditioned retrieval; ``Verif.'' denotes verifier-triggered fallback; ``CF'' denotes counterfactual minimal edits. \textbf{Bold} indicates best performance.}
\label{tab:ablation}
\small
\setlength{\tabcolsep}{4pt} 
\renewcommand{\arraystretch}{1.2}

\begin{tabular}{l ccccc cccccc cccccc}
\toprule
\multirow{2.5}{*}{\textbf{Variant}} & \multicolumn{5}{c}{\textbf{Modules}} & \multicolumn{6}{c}{\textbf{GPT-4 Backbone}} & \multicolumn{6}{c}{\textbf{LLaMA 3--8B Backbone}} \\
\cmidrule(lr){2-6} \cmidrule(lr){7-12} \cmidrule(lr){13-18}
 & Lat & Graph & RAG & Verif. & CF & CRC & SRS & Gnd. & HR $\downarrow$ & Prec & F1 & CRC & SRS & Gnd. & HR $\downarrow$ & Prec & F1 \\
\midrule

A0: Latent Baseline
& \checkmark & & & & 
& 0.73 & 0.67 & 0.64 & 0.13 & 0.72 & 0.74 
& 0.68 & 0.63 & 0.60 & 0.16 & 0.68 & 0.70 \\

A1: + Causal Graphs
& \checkmark & \checkmark & & & 
& 0.83 & 0.72 & 0.69 & 0.10 & 0.75 & 0.78 
& 0.78 & 0.69 & 0.66 & 0.13 & 0.72 & 0.75 \\

A2: + RAG Injection
& \checkmark & \checkmark & \checkmark & & 
& 0.91 & 0.76 & 0.71 & 0.08 & 0.76 & 0.79 
& 0.85 & 0.72 & 0.68 & 0.11 & 0.74 & 0.76 \\

A3: + Verifier Fallback
& \checkmark & \checkmark & \checkmark & \checkmark & 
& 0.96 & 0.82 & 0.75 & 0.05 & 0.79 & 0.83 
& 0.90 & 0.78 & 0.71 & 0.08 & 0.76 & 0.80 \\

\rowcolor{cHighlight}
A4: + Counterfactual (Full)
& \checkmark & \checkmark & \checkmark & \checkmark & \checkmark 
& \textbf{0.97} & \textbf{0.84} & \textbf{0.77} & \textbf{0.04} & \textbf{0.81} & \textbf{0.85} 
& \textbf{0.91} & \textbf{0.80} & \textbf{0.73} & \textbf{0.07} & \textbf{0.78} & \textbf{0.82} \\

\bottomrule
\end{tabular}
\end{table*}

\begin{figure*}[t]
\centering
\begin{subfigure}[t]{0.49\textwidth}
  \centering
  \includegraphics[width=\textwidth]{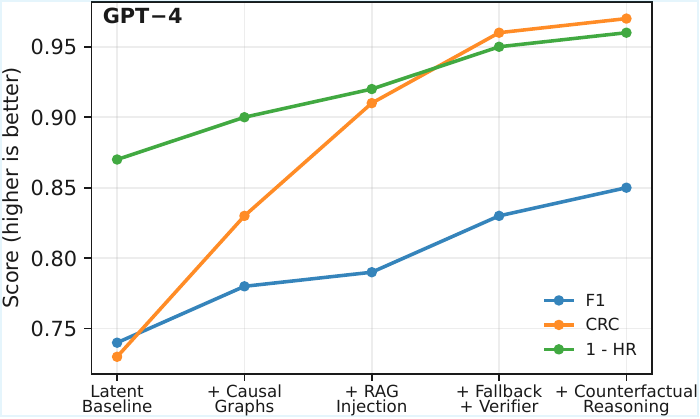}
  \caption{GPT-4}
  \label{fig:ablation_curve_gpt4}
\end{subfigure}\hfill
\begin{subfigure}[t]{0.49\textwidth}
  \centering
  \includegraphics[width=\textwidth]{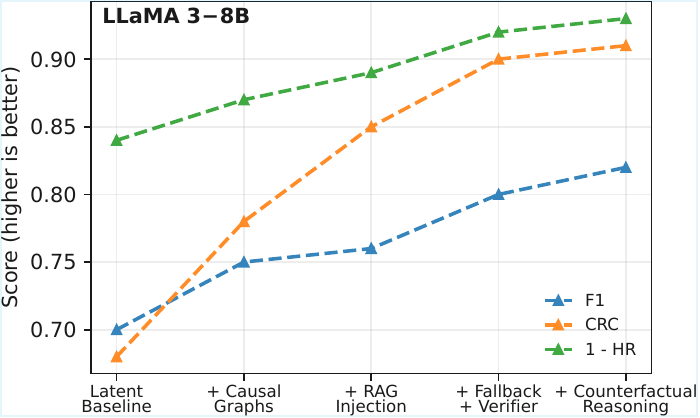}
  \caption{LLaMA 3--8B}
  \label{fig:ablation_curve_llama}
\end{subfigure}
\caption{\textbf{Incremental ablation curves for CARE-ECG.} Each panel visualizes how key metrics evolve as modules are progressively added (A0$\rightarrow$A4 in Table~\ref{tab:ablation}). The curves provide a compact view of the monotonic improvements and the stage at which each metric changes most strongly.}
\label{fig:ablation_curves}
\end{figure*}

\subsection{Ablation study: isolating module contributions and extracting insights}
While the aggregate results show consistent gains, they do not explain \emph{which} CARE-ECG components are responsible for improvements in each metric group. We therefore perform a staged ablation on Expert-ECG-QA, progressively enabling CARE-ECG mechanisms while holding the backbone and decoding settings constant. Each variant corresponds to a clear functional change:
\textbf{A0} uses only latent biomarkers and direct prompting;
\textbf{A1} adds Bayesian causal graph inference over $\tilde{z}$;
\textbf{A2} adds causal-conditioned retrieval;
\textbf{A3} adds verifier-triggered fallback; and
\textbf{A4} adds counterfactual minimal edits.

Table~\ref{tab:ablation} yields three core insights that are consistent across GPT-4 and LLaMA 3--8B. First, adding \textbf{causal graphs} (A0$\rightarrow$A1) produces a substantial increase in CRC (GPT-4: 0.73$\rightarrow$0.83; LLaMA: 0.68$\rightarrow$0.78) while also improving precision and F1 (GPT-4 F1: 0.74$\rightarrow$0.78; LLaMA F1: 0.70$\rightarrow$0.75). This suggests that graph-based inference does more than provide a probability distribution: it changes the \emph{structure} of reasoning by encouraging explanations to mention the variables that actually drive inference.

\begin{figure}[t]
\centering
\includegraphics[width=\columnwidth]{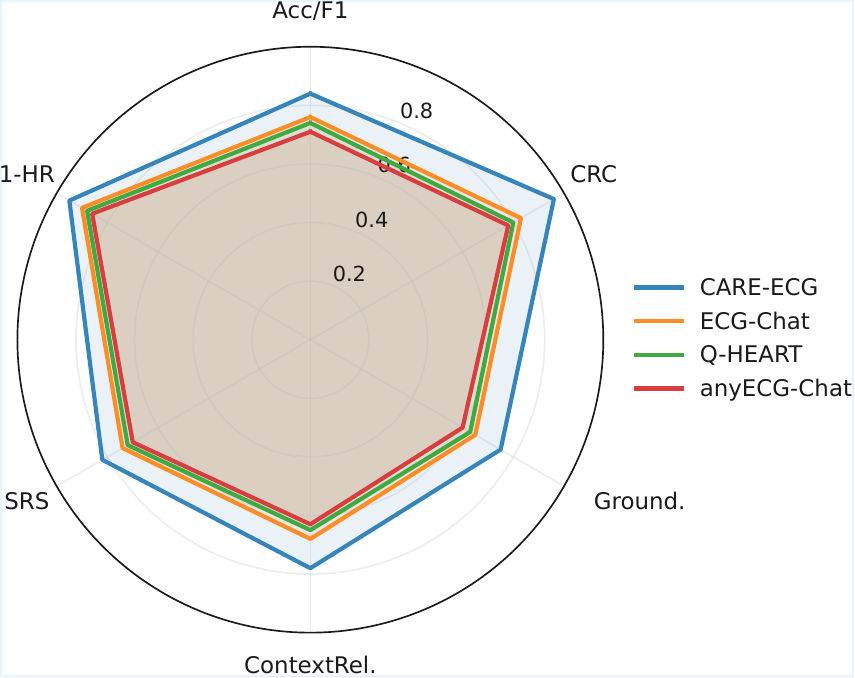}
\caption{\textbf{Radar-style summary of multi-metric trade-offs.} The radar view summarizes the consistent advantage of CARE-ECG across complementary reliability dimensions.}
\label{fig:radar_summary}
\end{figure}

Second, adding \textbf{causal-conditioned retrieval} (A1$\rightarrow$A2) further increases CRC and SRS while improving groundedness (GPT-4: 0.69$\rightarrow$0.71; LLaMA: 0.66$\rightarrow$0.68). This supports the intended division of labor: the graph identifies \emph{which physiological factors are salient}, while retrieval supplies \emph{the clinical criteria and guideline evidence} that justify them. Notably, the gains are not limited to groundedness; CRC and SRS also rise, indicating that retrieval does not merely add citations but helps anchor the explanation content to causal drivers.

Third, the \textbf{verifier + fallback} module is the dominant hallucination suppressor (A2$\rightarrow$A3). HR drops sharply for both backbones (GPT-4: 0.08$\rightarrow$0.05; LLaMA: 0.11$\rightarrow$0.08), while groundedness and SRS increase. This indicates that the safety layer is not a blunt constraint that reduces content; instead, it redirects generation toward evidence-consistent explanations when risk is detected. Finally, adding \textbf{counterfactual reasoning} (A3$\rightarrow$A4) yields further improvements across CRC/SRS/groundedness and reduces HR again (GPT-4: 0.05$\rightarrow$0.04; LLaMA: 0.08$\rightarrow$0.07). This suggests that minimal-edit counterfactuals strengthen mechanistic consistency: explanations must remain stable under small evidence perturbations, discouraging brittle correlations and encouraging causal narratives that match the system’s probabilistic sensitivity.

Fig.~\ref{fig:ablation_curves} complements Table~\ref{tab:ablation} by visualizing the same staged variants as a trajectory rather than a static table. This view makes two patterns easier to see. (i) \emph{Metric-specific inflection points:} CRC rises most sharply when causal graphs are introduced (A0$\rightarrow$A1), consistent with the interpretation that graph inference primarily strengthens causal anchoring, while HR drops most sharply when verifier-triggered fallback is added (A2$\rightarrow$A3), reflecting its role as the primary hallucination suppressor. (ii) \emph{Monotonicity across backbones:} both GPT-4 and LLaMA 3--8B exhibit the same qualitative trajectory, supporting the claim that improvements arise from pipeline structure rather than backbone-specific memorized medical knowledge.

Taken together, the ablation study supports the claim that CARE-ECG’s gains are not attributable to a single heuristic. Causal graphs improve causal alignment and structured correctness; retrieval improves evidence binding; verifier/fallback reduces hallucination; and counterfactual reasoning further strengthens mechanistic consistency. These modules contribute in complementary ways that align with the metric groups in Tables~\ref{tab:answer_quality}--\ref{tab:hallucination_factuality}, yielding a coherent picture: CARE-ECG is more correct, more faithful, and more robust.

Finally, Fig.~\ref{fig:radar_summary} provides a compact multi-metric summary that is aligned with the three reliability questions (Q1--Q3). While the tables report each metric precisely, the radar perspective highlights a practical takeaway for clinical deployment: improvements are not isolated to one axis (e.g., correctness alone), but appear jointly across correctness, causal/evidence alignment, and robustness. Importantly, this visualization is consistent with the numerical dominance shown in Tables~\ref{tab:answer_quality}--\ref{tab:hallucination_factuality} and the staged improvements in Table~\ref{tab:ablation}, reinforcing that CARE-ECG’s gains come from complementary modules rather than from trading off one reliability property against another.

\subsection{Summary}
Across three datasets and two backbones, CARE-ECG consistently improves answer quality (Table~\ref{tab:answer_quality}), causal/retrieval alignment (Table~\ref{tab:causal_retrieval}), and hallucination robustness (Table~\ref{tab:hallucination_factuality}). The ablation study (Table~\ref{tab:ablation}) provides mechanistic evidence for why these gains emerge: the causal structure constrains reasoning, retrieval anchors explanations to explicit evidence, and verification controls hallucination without sacrificing semantic relevance. Collectively, these results support CARE-ECG as a clinically oriented ECG--LLM framework that advances beyond fluent generation toward auditable and causally grounded interpretation.

\section{Discussion}

CARE-ECG advances ECG--LLM reasoning by introducing an explicit, audit-able intermediate state---the discretized latent biomarker regime $\tilde{z}$ and causal posterior $P(y\mid\tilde{z})$---that constrains both retrieval and explanation. Empirically, this structure helps align generated rationales with the model’s own drivers (higher CRC) and with retrieved clinical evidence (higher groundedness/SRS), while reducing unsupported waveform claims (lower HR). Beyond performance, the key advantage is \emph{traceability}: the system can expose which latent factors and causal pathways contribute to a diagnosis and how small edits would alter posterior risk, which is closer to how clinicians justify ECG decisions than direct waveform-to-text mapping.

However, CARE-ECG’s causal claims must be interpreted conservatively. The learned graph is constructed over representations extracted from observational data, and edges may reflect predictive dependencies, discretization artifacts, or dataset-specific confounding rather than true interventional causality. Similarly, counterfactual “minimal edits” are best viewed as \emph{local sensitivity analyses under the model}, not as actionable physiological interventions. Another limitation is that latent factors are not guaranteed to correspond cleanly to standard measurements (e.g., QTc, PR, QRS); dimensions can entangle morphology with noise, lead placement, or heart rate effects, and quantile binning may blur clinically meaningful threshold boundaries.

These limitations motivate several future directions. First, we will improve causal reliability by incorporating interventional or controlled perturbation supervision (e.g., synthetic ECG edits, protocol-based morphology changes) and by anchoring subsets of $z$ to measurable interval/morphology descriptors to enhance interpretability. Second, we will strengthen evaluation beyond overlap-style metrics by adding waveform-criterion checkers (explicit interval/ST computations) and clinician-graded evidence attribution, especially under distribution shifts (wearables vs.\ hospital ECGs) and adversarial expert questioning. Finally, we will optimize the pipeline for deployment by distilling evidence-grounded behavior into lighter models and by exploring privacy-aware sharing of intermediate traces, ensuring auditability without exposing sensitive patient-specific representations.

\section{Conclusion}
We introduced CARE-ECG, a causally structured ECG--language reasoning framework that encodes multi-lead ECGs into temporally organized latent biomarkers, performs Bayesian causal inference with counterfactual sensitivity analysis, and generates evidence-grounded explanations via causal-conditioned retrieval with verifier-triggered fallback. Across PTB-XL, MIMIC-IV ECG, and Expert-ECG-QA, CARE-ECG improves clinical correctness, strengthens causal/evidence alignment, and reduces hallucinations under both GPT-4 and LLaMA backbones, while exposing intermediate reasoning states for audit and verification. These results highlight that explicit causal structure and evidence control are key ingredients for building safer and more clinically trustworthy ECG--LLM decision-support systems.

\bibliographystyle{ACM-Reference-Format}
\bibliography{refs}

@article{wang2025healthq,
  title={Healthq: Unveiling questioning capabilities of llm chains in healthcare conversations},
  author={Wang, Ziyu and Li, Hao and Huang, Di and Kim, Hye-Sung and Shin, Chae-Won and Rahmani, Amir M},
  journal={Smart Health},
  pages={100570},
  year={2025},
  publisher={Elsevier}
}

@inproceedings{wang2024ecg,
  title={Ecg unveiled: Analysis of client re-identification risks in real-world ecg datasets},
  author={Wang, Ziyu and Kanduri, Anil and Aqajari, Seyed Amir Hossein and Jafarlou, Salar and Mousavi, Sanaz R and Liljeberg, Pasi and Malik, Shaista and Rahmani, Amir M},
  booktitle={2024 IEEE 20th International Conference on Body Sensor Networks (BSN)},
  pages={1--4},
  year={2024},
  organization={IEEE}
}

@article{ong2024medical,
  title={Medical ethics of large language models in medicine},
  author={Ong, Jasmine Chiat Ling and Chang, Shelley Yin-Hsi and William, Wasswa and Butte, Atul J and Shah, Nigam H and Chew, Lita Sui Tjien and Liu, Nan and Doshi-Velez, Finale and Lu, Wei and Savulescu, Julian and others},
  journal={NEJM AI},
  volume={1},
  number={7},
  pages={AIra2400038},
  year={2024},
  publisher={Massachusetts Medical Society}
}

@article{wang2025linkage,
  title={Linkage Attacks Expose Identity Risks in Public ECG Data Sharing},
  author={Wang, Ziyu and Khatibi, Elahe and Firouzi, Farshad and Mousavi, Sanaz Rahimi and Chakrabarty, Krishnendu and Rahmani, Amir M},
  journal={arXiv preprint arXiv:2508.15850},
  year={2025}
}

@article{wang2025medcot,
  title={MedCoT-RAG: Causal Chain-of-Thought RAG for Medical Question Answering},
  author={Wang, Ziyu and Khatibi, Elahe and Rahmani, Amir M},
  journal={arXiv preprint arXiv:2508.15849},
  year={2025}
}

@article{khatibi2025cdf,
  title={CDF-RAG: Causal Dynamic Feedback for Adaptive Retrieval-Augmented Generation},
  author={Khatibi, Elahe and Wang, Ziyu and Rahmani, Amir M},
  journal={arXiv preprint arXiv:2504.12560},
  year={2025}
}

@article{aqajari2024enhancing,
  title={Enhancing performance and user engagement in everyday stress monitoring: A context-aware active reinforcement learning approach},
  author={Aqajari, Seyed Amir Hossein and Wang, Ziyu and Tazarv, Ali and Labbaf, Sina and Jafarlou, Salar and Nguyen, Brenda and Dutt, Nikil and Levorato, Marco and Rahmani, Amir M},
  journal={arXiv preprint arXiv:2407.08215},
  year={2024}
}

@inproceedings{alikhani2024seal,
  title={SEAL: Sensing efficient active learning on wearables through context-awareness},
  author={Alikhani, Hamidreza and Wang, Ziyu and Kanduri, Anil and Lilieberg, Pasi and Rahmani, Amir M and Dutt, Nikil},
  booktitle={2024 Design, Automation \& Test in Europe Conference \& Exhibition (DATE)},
  pages={1--2},
  year={2024},
  organization={IEEE}
}

@inproceedings{alikhani2024ea,
  title={EA\^{} 2: Energy Efficient Adaptive Active Learning for Smart Wearables},
  author={Alikhani, Hamidreza and Wang, Ziyu and Kanduri, Anil and Liljeberg, Pasi and Rahmani, Amir M and Dutt, Nikil},
  booktitle={Proceedings of the 29th ACM/IEEE international symposium on low power electronics and design},
  pages={1--6},
  year={2024}
}

@article{yu2022ai,
  title={AI-based stroke disease prediction system using ECG and PPG bio-signals},
  author={Yu, Jaehak and Park, Sejin and Kwon, Soon-Hyun and Cho, Kang-Hee and Lee, Hansung},
  journal={Ieee Access},
  volume={10},
  pages={43623--43638},
  year={2022},
  publisher={IEEE}
}

@article{murat2020application,
  title={Application of deep learning techniques for heartbeats detection using ECG signals-analysis and review},
  author={Murat, Fatma and Yildirim, Ozal and Talo, Muhammed and Baloglu, Ulas Baran and Demir, Yakup and Acharya, U Rajendra},
  journal={Computers in biology and medicine},
  volume={120},
  pages={103726},
  year={2020},
  publisher={Elsevier}
}

@inproceedings{yu2023zero,
  title={Zero-shot ECG diagnosis with large language models and retrieval-augmented generation},
  author={Yu, Han and Guo, Peikun and Sano, Akane},
  booktitle={Machine learning for health (ML4H)},
  pages={650--663},
  year={2023},
  organization={PMLR}
}

@inproceedings{wang2025llm,
  title={Llm-rg4: Flexible and factual radiology report generation across diverse input contexts},
  author={Wang, Zhuhao and Sun, Yihua and Li, Zihan and Yang, Xuan and Chen, Fang and Liao, Hongen},
  booktitle={Proceedings of the AAAI Conference on Artificial Intelligence},
  volume={39},
  number={8},
  pages={8250--8258},
  year={2025}
}

@article{singhal2025toward,
  title={Toward expert-level medical question answering with large language models},
  author={Singhal, Karan and Tu, Tao and Gottweis, Juraj and Sayres, Rory and Wulczyn, Ellery and Amin, Mohamed and Hou, Le and Clark, Kevin and Pfohl, Stephen R and Cole-Lewis, Heather and others},
  journal={Nature Medicine},
  pages={1--8},
  year={2025},
  publisher={Nature Publishing Group US New York}
}

@article{kim2025medical,
  title={Medical hallucinations in foundation models and their impact on healthcare},
  author={Kim, Yubin and Jeong, Hyewon and Chen, Shan and Li, Shuyue Stella and Lu, Mingyu and Alhamoud, Kumail and Mun, Jimin and Grau, Cristina and Jung, Minseok and Gameiro, Rodrigo and others},
  journal={arXiv preprint arXiv:2503.05777},
  year={2025}
}

@article{gow2023mimic,
  title={MIMIC-IV-ECG: Diagnostic Electrocardiogram Matched Subset},
  author={Gow, Brian and Pollard, Tom and Nathanson, Larry A and Johnson, Alistair and Moody, Benjamin and Fernandes, Chrystinne and Greenbaum, Nathaniel and Waks, Jonathan W and Eslami, Parastou and Carbonati, Tanner and others},
  journal={Type: dataset},
  volume={6},
  pages={13--14},
  year={2023}
}

@article{wagner2020ptb,
  title={PTB-XL, a large publicly available electrocardiography dataset},
  author={Wagner, Patrick and Strodthoff, Nils and Bousseljot, Ralf-Dieter and Kreiseler, Dieter and Lunze, Fatima I and Samek, Wojciech and Schaeffter, Tobias},
  journal={Scientific data},
  volume={7},
  number={1},
  pages={1--15},
  year={2020},
  publisher={Nature Publishing Group}
}

@article{yang2020causalvae,
  title={Causalvae: Structured causal disentanglement in variational autoencoder},
  author={Yang, Mengyue and Liu, Furui and Chen, Zhitang and Shen, Xinwei and Hao, Jianye and Wang, Jun},
  journal={arXiv preprint arXiv:2004.08697},
  year={2020}
}

@article{gu2021efficiently,
  title={Efficiently modeling long sequences with structured state spaces},
  author={Gu, Albert and Goel, Karan and R{\'e}, Christopher},
  journal={arXiv preprint arXiv:2111.00396},
  year={2021}
}

@article{wang2025ecg,
  title={ECG-Expert-QA: A Benchmark for Evaluating Medical Large Language Models in Heart Disease Diagnosis},
  author={Wang, Xu and Kang, Jiaju and Han, Puyu and Zhao, Yubao and Liu, Qian and He, Liwenfei and Zhang, Lingqiong and Dai, Lingyun and Wang, Yongcheng and Tao, Jie},
  journal={arXiv preprint arXiv:2502.17475},
  year={2025}
}

@article{singhal2024medpalm2,
  title={Toward expert-level medical question answering with large language models},
  author={Singhal, Karan and Tu, Tao and Gottweis, Jurgen and Sayres, Rory and Wulczyn, Ellery and Hou, Le and Clark, Kristy and Pfohl, Stephen and Cole-Lewis, Heather and Neal, David and others},
  journal={Nature Medicine},
  year={2024},
  publisher={Nature Publishing Group},
  doi={10.1038/s41591-024-03423-7}
}

@article{wei2022chain,
  title={Chain-of-thought prompting elicits reasoning in large language models},
  author={Wei, Jason and Wang, Xuezhi and Schuurmans, Dale and Bosma, Maarten and Xia, Fei and Chi, Ed and Le, Quoc V and Zhou, Denny and others},
  journal={Advances in neural information processing systems},
  volume={35},
  pages={24824--24837},
  year={2022}
}

@article{shool2025systematic,
  title={A systematic review of large language model (LLM) evaluations in clinical medicine},
  author={Shool, Sina and Adimi, Sara and Saboori Amleshi, Reza and Bitaraf, Ehsan and Golpira, Reza and Tara, Mahmood},
  journal={BMC Medical Informatics and Decision Making},
  volume={25},
  number={1},
  pages={117},
  year={2025},
  publisher={Springer}
}

@article{ebrahimi2020review,
  title={A review on deep learning methods for ECG arrhythmia classification},
  author={Ebrahimi, Zahra and Loni, Mohammad and Daneshtalab, Masoud and Gharehbaghi, Arash},
  journal={Expert Systems with Applications: X},
  volume={7},
  pages={100033},
  year={2020},
  publisher={Elsevier}
}

@article{hannun2019cardiologist,
  title={Cardiologist-level arrhythmia detection and classification in ambulatory electrocardiograms using a deep neural network},
  author={Hannun, Awni Y and Rajpurkar, Pranav and Haghpanahi, Masoumeh and Tison, Geoffrey H and Bourn, Codie and Turakhia, Mintu P and Ng, Andrew Y},
  journal={Nature medicine},
  volume={25},
  number={1},
  pages={65--69},
  year={2019},
  publisher={Nature Publishing Group US New York}
}

@article{ribeiro2020automatic,
  title={Automatic diagnosis of the 12-lead ECG using a deep neural network},
  author={Ribeiro, Ant{\^o}nio H and Ribeiro, Manoel Horta and Paix{\~a}o, Gabriela MM and Oliveira, Derick M and Gomes, Paulo R and Canazart, J{\'e}ssica A and Ferreira, Milton PS and Andersson, Carl R and Macfarlane, Peter W and Meira Jr, Wagner and others},
  journal={Nature communications},
  volume={11},
  number={1},
  pages={1760},
  year={2020},
  publisher={Nature Publishing Group UK London}
}

@article{mckeen2025ecg,
  title={Ecg-fm: An open electrocardiogram foundation model},
  author={McKeen, Kaden and Masood, Sameer and Toma, Augustin and Rubin, Barry and Wang, Bo},
  journal={JAMIA open},
  volume={8},
  number={5},
  pages={ooaf122},
  year={2025},
  publisher={Oxford University Press}
}

@book{asher1976causal,
  title={Causal modeling},
  author={Asher, Herbert B},
  volume={3},
  year={1976},
  publisher={Sage}
}

@article{perzhilla2025situ,
  title={In-situ dehydration monitoring via a stable diffusion-aided single-lead ecg iomt: Ml/dl models shine while llms hallucinate},
  author={Perzhilla, Levina and Siyoucef, Soumia and Al-Aslani, Rose and Rahman, Muhammad Mahboob Ur and Al-Naffouri, Tareq Y},
  journal={IEEE Internet of Things Journal},
  year={2025},
  publisher={IEEE}
}

@inproceedings{caitowards,
  title={Towards Generalizable Multimodal ECG Representation Learning with LLM-extracted Clinical Entities},
  author={Cai, Mingsheng and Jiang, Jiuming and Huang, Wenhao and Liu, Che and Arcucci, Rossella},
  booktitle={1st ICML Workshop on Foundation Models for Structured Data}
}

@inproceedings{yang2021causalvae,
  title={Causalvae: Disentangled representation learning via neural structural causal models},
  author={Yang, Mengyue and Liu, Furui and Chen, Zhitang and Shen, Xinwei and Hao, Jianye and Wang, Jun},
  booktitle={Proceedings of the IEEE/CVF conference on computer vision and pattern recognition},
  pages={9593--9602},
  year={2021}
}

@article{dao2024transformers,
  title={Transformers are ssms: Generalized models and efficient algorithms through structured state space duality},
  author={Dao, Tri and Gu, Albert},
  journal={arXiv preprint arXiv:2405.21060},
  year={2024}
}
\end{document}